\documentclass[sigconf]{acmart}

\copyrightyear{2023}
\acmYear{2023}
\setcopyright{rightsretained}
\acmConference[KDD '23]{Proceedings of the 29th ACM SIGKDD Conference on Knowledge Discovery and Data Mining}{August 6--10, 2023}{Long Beach, CA, USA}
\acmBooktitle{Proceedings of the 29th ACM SIGKDD Conference on Knowledge Discovery and Data Mining (KDD '23), August 6--10, 2023, Long Beach, CA, USA}
\acmDOI{10.1145/3580305.3599929}
\acmISBN{979-8-4007-0103-0/23/08}

\makeatletter

\usepackage{amsmath}
\usepackage{amsfonts}
\usepackage{subcaption}
\usepackage{graphicx}
\usepackage{booktabs}
\usepackage{url}
\usepackage{tabularx}
\usepackage{bigstrut}
\usepackage{multirow}
\usepackage{color}
\usepackage{xcolor}
\usepackage{verbatimbox}
\usepackage{enumitem}
\usepackage{xspace}
\usepackage{bbding}
\usepackage{algorithm} 
\usepackage{algpseudocode}
\usepackage{enumitem}
\usepackage{color}
\usepackage{pifont}
\newcommand{\xmark}{\ding{55}}
\newcommand{\cmark}{\ding{51}}%

\begin{document}

\title{VRDU: A Benchmark for Visually-rich Document Understanding}

\author{Zilong Wang}
\authornote{This work was completed while the author was working as an intern at Google Research.}
\affiliation{
  \institution{University of California, San Diego}
  \city{La Jolla}
  \country{USA}
}
\email{zlwang@ucsd.edu}

\author{Yichao Zhou}
\affiliation{%
  \institution{Google}
  \city{Mountain View}
  \country{USA}
}
\email{yichaojoey@google.com}

\author{Wei Wei}
\affiliation{
  \institution{Google}
  \city{Sunnyvale}
  \country{USA}
}
\email{wewei@google.com}

\author{Chen-Yu Lee}
\affiliation{
  \institution{Google}
  \city{Sunnyvale}
  \country{USA}
}
\email{chenyulee@google.com}

\author{Sandeep Tata}
\affiliation{
  \institution{Google}
  \city{Mountain View}
  \country{USA}
}
\email{tata@google.com}




\newcommand{\zilong}[1]{\textcolor{blue}{#1}}

\begin{abstract}
  Understanding visually-rich business documents to extract structured data and automate business workflows has been receiving attention both in academia and industry.  Although recent multi-modal language models have achieved impressive results, we find that existing benchmarks do not reflect the complexity of real documents seen in industry. In this work, we identify the desiderata for a more comprehensive benchmark and propose one we call Visually Rich Document Understanding (VRDU). VRDU contains two datasets that represent several challenges: rich schema including diverse data types as well as hierarchical entities, complex templates including tables and multi-column layouts, and diversity of different layouts (templates) within a single document type. We design few-shot and conventional experiment settings along with a carefully designed matching algorithm to evaluate extraction results. We report the performance of strong baselines and offer three observations: (1) generalizing to new document templates is still very challenging, (2) few-shot performance has a lot of headroom, and (3) models struggle with hierarchical fields such as line-items in an invoice. We plan to open source the benchmark and the evaluation toolkit. We hope this helps the community make progress on these challenging tasks in extracting structured data from visually rich documents.

\end{abstract}



\maketitle

\section{Introduction}
Visually-rich documents, such as forms, receipts, invoices, are ubiquitous in various business workflows. Distinct from plain text documents, visually-rich documents have layout information that is critical to the understanding of documents. Given the potential to automate business workflows across procurement, banking, insurance, retail lending, healthcare, etc., understanding these documents, and in particular extracting structured objects from them has recently received a lot of attention from both industry and academia~\cite{li2020docbank,zhang2020trie,powalski2021going,appalaraju2021docformer,garncarek2021lambert,biten2022latr}.

While tasks such as classification~\cite{Harley2015EvaluationOD} and Visual-QA~\cite{mathew2021docvqa} have been posed to study the understanding of such documents, in this paper, we focus on the task of extracting structured information. Optical character recognition engines (OCR) are typically used to extract the textual content and the bounding boxes of each of the words from the documents. Existing models rely on language models with multi-modal features to solve the task, where features from textual contents, images, and structural templates are jointly encoded through self-supervised training ~\cite{xu2020layoutlmv2,xu2020layoutlm,appalaraju2021docformer,garncarek2021lambert,lee2022formnet,powalski2021going}. Although recent models achieved impressive results ~\cite{jaume2019,park2019cord,huang2019icdar2019,stanislawek2021kleister}, we argue that existing benchmarks do not reflect the challenges encountered in practice, such as having to generalize to unseen templates, complex target schema, hierarchical entities, and small training sets. 

\begin{figure*}[t]
    \centering
    \includegraphics[width=0.90\linewidth]{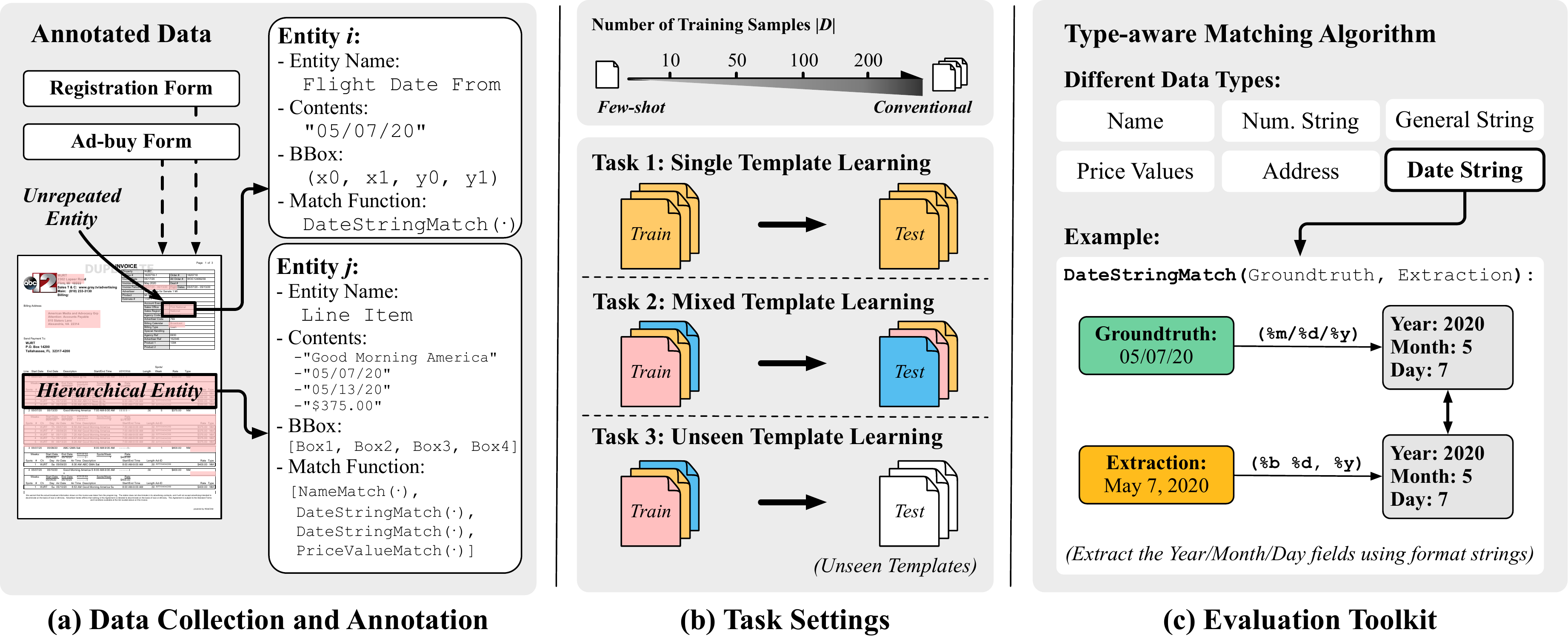}
    \caption{Overview of the VRDU benchmark: (a) high-quality annotation of rich labeling schema; (b) tasks of different difficulty levels and different number of training samples; (c) type-aware matching algorithm for entities of different data types.}
    \label{fig:benchmark}
\end{figure*}

We identify five desiderata (Section \ref{sec:desiderata}) for benchmarks on this topic based on our observations of drawbacks of existing datasets. First, most existing benchmarks suffer from the fact that they lack richness in labeling schema~\cite{jaume2019,huang2019icdar2019,stanislawek2021kleister}. Entities are roughly considered as simple text strings while practical document types have a variety of types like numerical IDs, dates, addresses, currency amounts, etc. Further, real-world docs frequently have hierarchical and repeated fields like componentized addresses and line-items in invoices. Second, some benchmarks contain documents with limited layout complexity. Pages that are mostly organized in long paragraphs and sentences are more similar to plain text documents~\cite{stanislawek2021kleister} and are not helpful evaluating our understanding of visually-rich documents. Third, the documents in some benchmarks may share the same template~\cite{huang2019icdar2019}. This makes it trivial for the models to deal with these document by simply memorizing the structure even if the single template is complex. Next, existing datasets use different OCR engines~\cite{park2019cord,huang2019icdar2019}. The large variety of OCR engines make it hard to tell whether the improvements come from the advanced models or more accurate OCR results. Finally, some benchmarks only provide the textual contents for each entity without further annotating the specific tokens in the document that are involved in the entities~\cite{huang2019icdar2019,stanislawek2021kleister,svetlichnaya2020deepform}, which means the models cannot be supervised with the token-level annotation. While this seems minor, it is very difficult to re-construct the token-level annotation only with textual contents of entities since the same text (e.g. ``0.0'') may appear multiple times in the document but only one of them may correspond to the target entity. It is necessary to involve human annotators to fix the issue by relabeling the documents with precise token spans. Also note that most existing approaches on this topic are based on sequence labeling models~\cite{xu2020layoutlm,xu2020layoutlmv2,huang2022layoutlmv3,zhang2020trie,appalaraju2021docformer,garncarek2021lambert,huang2022layoutlmv3,wang2022lilt} that require token-level annotations to work.

Based on these observations, we propose a new benchmark, \textbf{VRDU}, for \textbf{V}isually-\textbf{R}ich \textbf{D}ocument \textbf{U}nderstanding task. VRDU is designed to reflect the challenges encountered in practice and eliminate the unnecessary factors affecting the research. We hope that this benchmark helps bridge the gap between academic research and practical scenarios to facilitate future study on this topic. As shown in Figure \ref{fig:benchmark}, we collected political ad-buy forms from the Federal Communications Commission (FCC)\footnote{\url{https://publicfiles.fcc.gov}} and registration forms from the Foreign Agents Registration Act (FARA)\footnote{\url{https://www.justice.gov}}, and constructed two datasets. We describe the annotated data, and the labeling protocol in Section \ref{sec:benchmark}.

Based on the two datasets, we then design three tasks of increasing difficulty. The tasks are designed to be similar to real applications. In Task 1 Single Template Learning, documents in the train and test sets are drawn from a single template. In Task 2 Mixed Template Learning, we increase the diversity of templates, but train and test sets for each document type are drawn from the same set of templates. In Task 3 Unseen Template Learning, the train and test sets are drawn from disjoint sets of templates to measure how well a model generalizes to unseen templates. Within each task, we compare the model performance with different number of training samples to understand the data efficiency for each approach. Finally, we evaluate the model performance with a type-aware match algorithm, where we use different matching functions for each entity according to its data type instead of simply using string matching when comparing the prediction results with the groundtruth. For example, when comparing numerical entities, we may want ``4'' and ``4.0'' to be considered equivalent, while for address fields, ``4, Main St.'' and ``40 Main St.'' ought not to be considered equivalent.

We report the performance of commonly-used baseline models, LayoutLM~\cite{xu2020layoutlm}, LayoutLMv2~\cite{xu2020layoutlmv2}, LayoutLMv3~\cite{huang2022layoutlmv3}, and FormNet~\cite{lee2022formnet} in each task. Our work is \emph{not} meant to be a comparison of these model architectures. Through our experiments, we highlight three areas of opportunity for all these models. First, while the models are great at extracting from new instances of documents with a layout that matches one seen during training (Task 1 Single Template Learning and Task 2 Mixed Template Learning), they do worse on new layouts (Task 3 Unseen Template Learning). Second, few-shot performance continues to be hard with substantial room for improvement. Third, extracting hierarchical or repeated entities is really challenging, and all models perform worse on this compared to simple fields. 

We summarize our contribution as follows.
\begin{itemize}[nosep, leftmargin=*]
    \item We identify desiderata for benchmarks in the visually-rich document understanding task, arguing that the current datasets do not meet these requirements. 
    \item We propose VRDU, a new comprehensive benchmark for visually-rich document understanding. We open-source the dataset with high-quality OCR results and annotations. We also define three tasks corresponding to different application scenarios, and open-source  an evaluation toolkit with a type-aware matching algorithm. The toolkit and dataset can be found at \url{https://github.com/google-research/google-research/tree/master/vrdu}.
    \item VRDU satisfies all of our proposed desiderata and reflects practical challenges in extracting structured data from visually rich documents. It bridges the gap between academic research and practical scenarios to facilitate future study on this topic.
    \item Through experiments on multiple commonly-used baseline models, we show that there is substantial room for progress on the tasks in VRDU with regard to template transfer learning, few-shot settings, and hierarchical entity extractions.
\end{itemize}
\section{Benchmark Desiderata}\label{sec:desiderata}
\label{section:desiderata}
We identify five key desiderata for a benchmark that reflects practical challenges in extracting structured data from visually rich documents. A benchmark on the visually-rich document understanding topic should involve rich schema, layout-rich documents, diverse template, high-quality OCR results, and token-level annotation.

\subsection{Rich Schema}
\begin{figure}[h]
    \centering
    \includegraphics[width=0.95\linewidth]{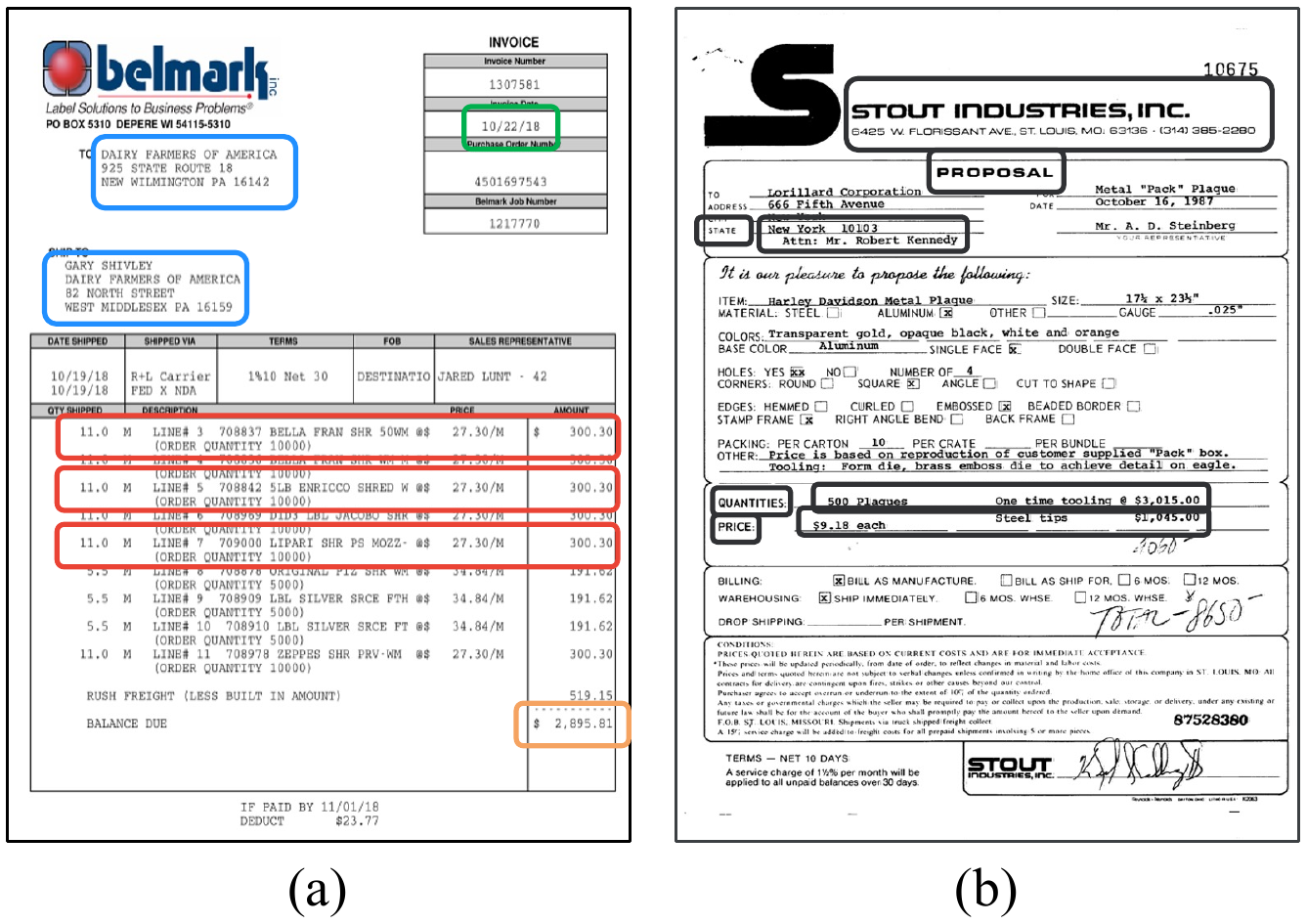}
    \caption{Examples of labeling schema:
    (a) specifies data types for each text fields, such as \textit{date strings}, \textit{address lines}, \textit{price values}, and \textit{hierarchical entities} (denoted with
    \colorbox[rgb]{0,0.69,0.31}{\textcolor[rgb]{0,0.69,0.31}{Bl}},
    \colorbox[rgb]{0.12,0.56,1.0}{\textcolor[rgb]{0.12,0.56,1.0}{Bl}},
    \colorbox[rgb]{0.96,0.64,0.38}{\textcolor[rgb]{0.96,0.64,0.38}{Bl}},
    \colorbox[rgb]{1,0,0}{\textcolor[rgb]{1,0,0}{Bl}}, 
    respectively);
    (b) treats all text fields as \textit{simple text strings} ignoring the specific data type (denoted with \colorbox[rgb]{0,0,0}{\textcolor[rgb]{0,0,0}{Bl}}).
    }
    \label{fig:rich_schema}
\end{figure}

The structured data we need to extract from in practice reflect a rich diversity of schemas. Entities extracted have various types such as numerical IDs, names, addresses, dates, currency amounts, etc. They can be required, optional, or repeated for a given document. In several cases, we also see hierarchical entities. For example, a US address field contains address lines, city, state, and zip code. A hierarchical entity is composed of all these components. Considering the heterogeneity of schema we encounter in practical settings, we believe a useful benchmark should reflect a rich schema. Contrast this with a dataset (see Figure~\ref{fig:rich_schema}) where the entities to be extracted are all treated as simple text strings named \emph{header}, \emph{question}, and \emph{answer}.

\subsection{Layout-rich Documents}
\begin{figure}[h]
    \centering
    \includegraphics[width=0.95\linewidth]{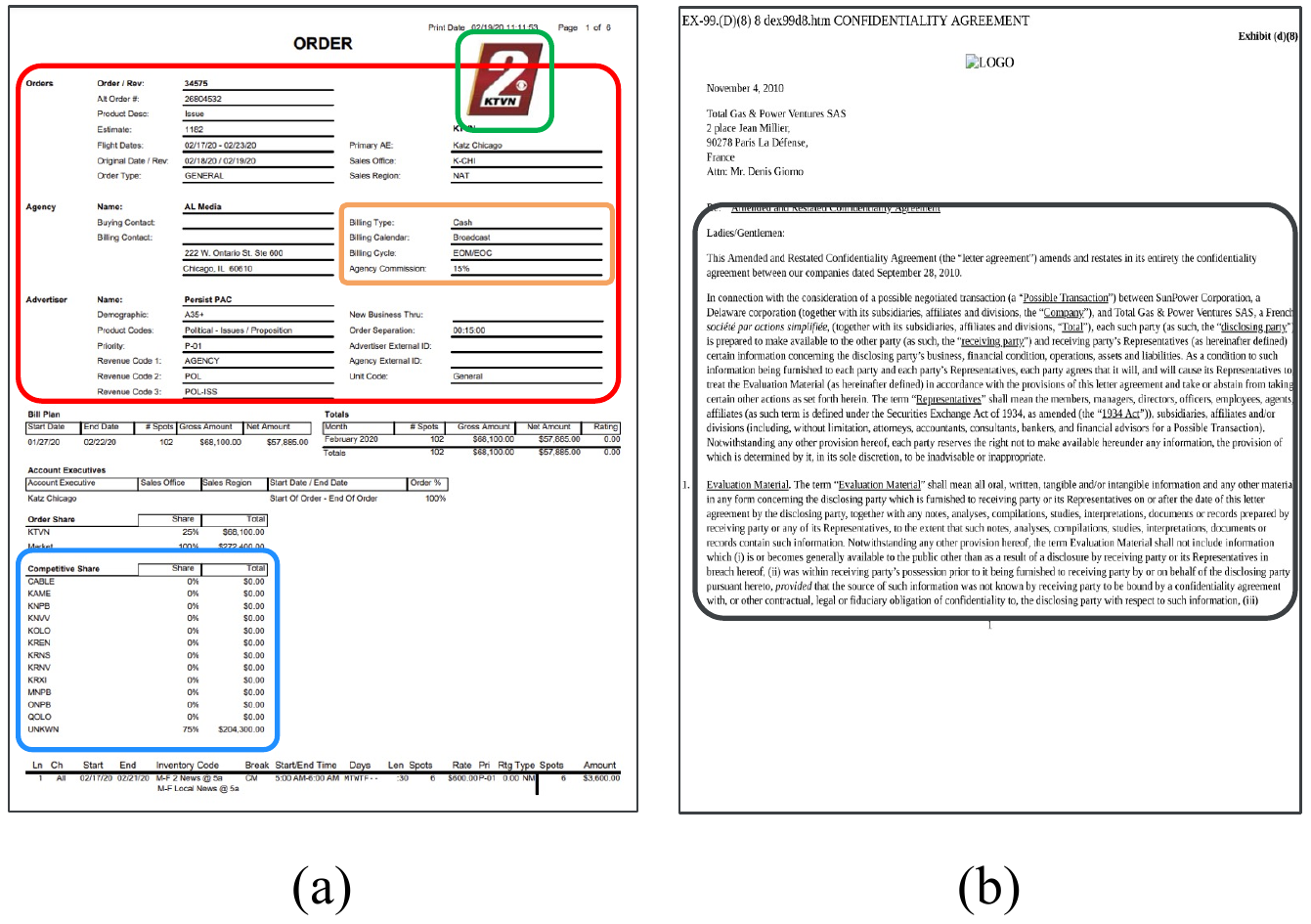}
    \caption{Examples of layout elements in the documents:
    (a) involves rich layout elements, such as \textit{images}, \textit{tables}, \textit{key-value pairs}, and \textit{multi-columns} (denoted with
    \colorbox[rgb]{0,0.69,0.31}{\textcolor[rgb]{0,0.69,0.31}{Bl}},
    \colorbox[rgb]{0.12,0.56,1.0}{\textcolor[rgb]{0.12,0.56,1.0}{Bl}},
    \colorbox[rgb]{0.96,0.64,0.38}{\textcolor[rgb]{0.96,0.64,0.38}{Bl}},
    \colorbox[rgb]{1,0,0}{\textcolor[rgb]{1,0,0}{Bl}}, 
    respectively);
    (b) largely contains natural language like \textit{paragraphs}, \textit{sentences}, \textit{chapters} (denoted with \colorbox[rgb]{0,0,0}{\textcolor[rgb]{0,0,0}{Bl}}).}
    \label{fig:layout_rich}
\end{figure}

The documents should have complex layout elements. Challenges in practical settings come from the fact that documents may contain tables, key-value pairs, switch between single-column and double-column layout, have varying font-sizes for different sections, include pictures with captions, and even footnotes. Contrast this with datasets where most documents are organized in sentences, paragraphs, and chapters with  section headers. Figure~\ref{fig:layout_rich} shows an example
of a document with rich layout and contrasts it with a more traditional document that is the focus of classic NLP literature on long inputs.


\subsection{Diverse Templates}
\begin{figure}[h]
    \centering
    \includegraphics[width=0.95\linewidth]{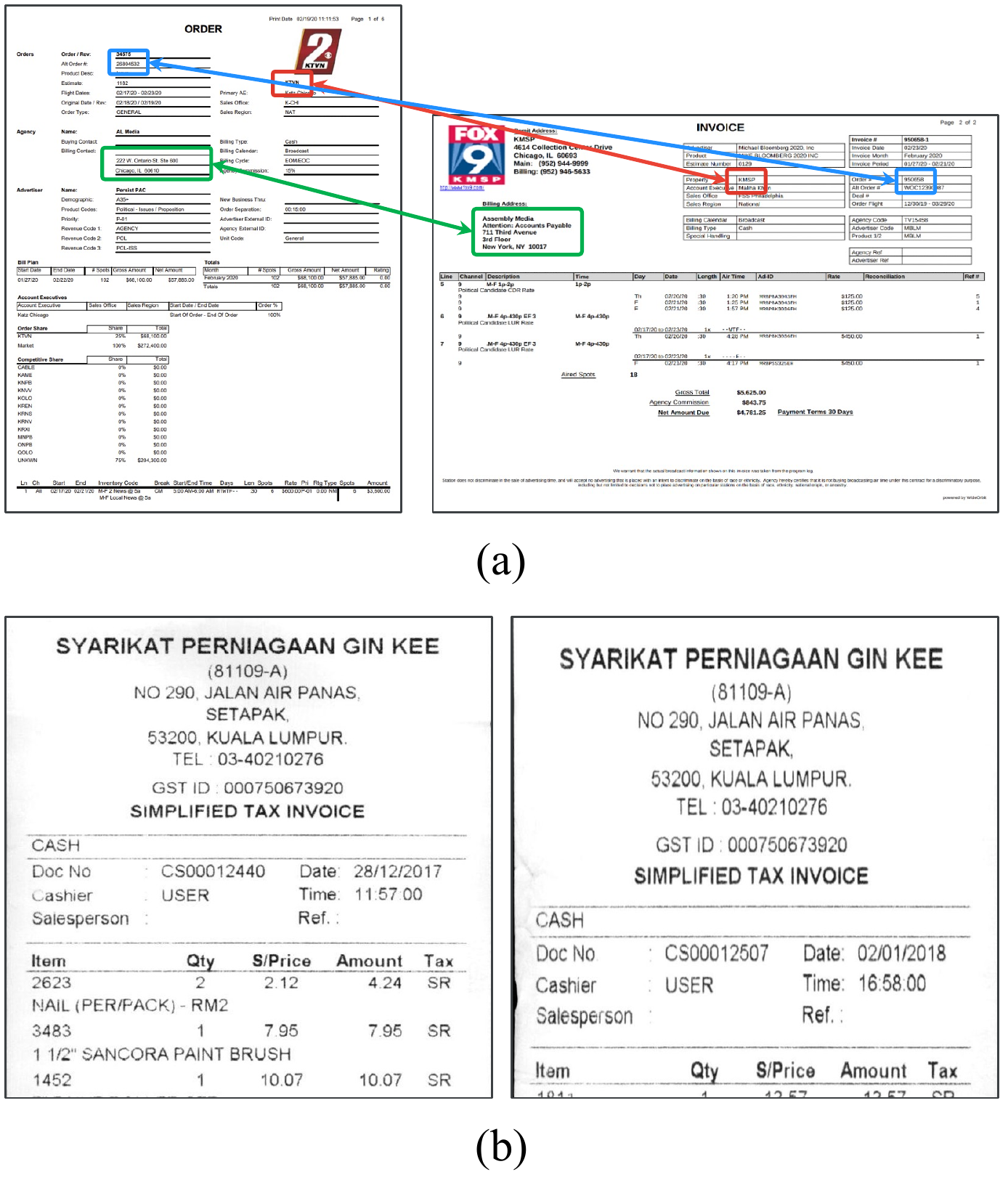}
    \caption{Examples of document templates:
    (a) two examples of the same document type with different templates (entities denoted with
    \colorbox[rgb]{0,0.69,0.31}{\textcolor[rgb]{0,0.69,0.31}{Bl}},
    \colorbox[rgb]{0.12,0.56,1.0}{\textcolor[rgb]{0.12,0.56,1.0}{Bl}},
    \colorbox[rgb]{1,0,0}{\textcolor[rgb]{1,0,0}{Bl}} for \textit{address}, \textit{contract ID}, and \textit{TV station name}, respectively);
    (b) example of different documents that share the same template.
    }
    \label{fig:diverse_template}
\end{figure}

A benchmark collection should involve different structural layouts or templates as shown in Figure~\ref{fig:diverse_template}. It is trivial to extract from a particular template by memorizing the structure. However, in practice one needs to be able to generalize to new templates. Consider, for instance, an invoice parser. If a company starts working with a new vendor (and enterprises routinely work with new vendors every year), a model that memorized the set of templates corresponding to existing vendors is likely to break since the new vendor may send invoices with a different template. In order to reflect this real-world requirement, a useful benchmark for extraction from visually-rich documents should have diverse templates and test a model's ability to generalize to unseen templates. 

\subsection{High-quality OCR Results}

Documents should have high-quality OCR results. Our aim with this benchmark is to focus on the VRDU task itself and we want to exclude the variability brought on by the choice of OCR engine. Existing benchmarks use different OCR engines, which makes the evaluation results inconsistent and the comparison unfair. It is confusing whether the performance improvements come from the more advanced model design or are simply because of more accurate OCR results. Therefore, a benchmark should use the same high-quality engine ensuring the quality of OCR is satisfactory and the choice of OCR engine is not a factor influencing the results when comparing the performance.

\begin{figure}
    \centering
    \includegraphics[width=0.95\linewidth]{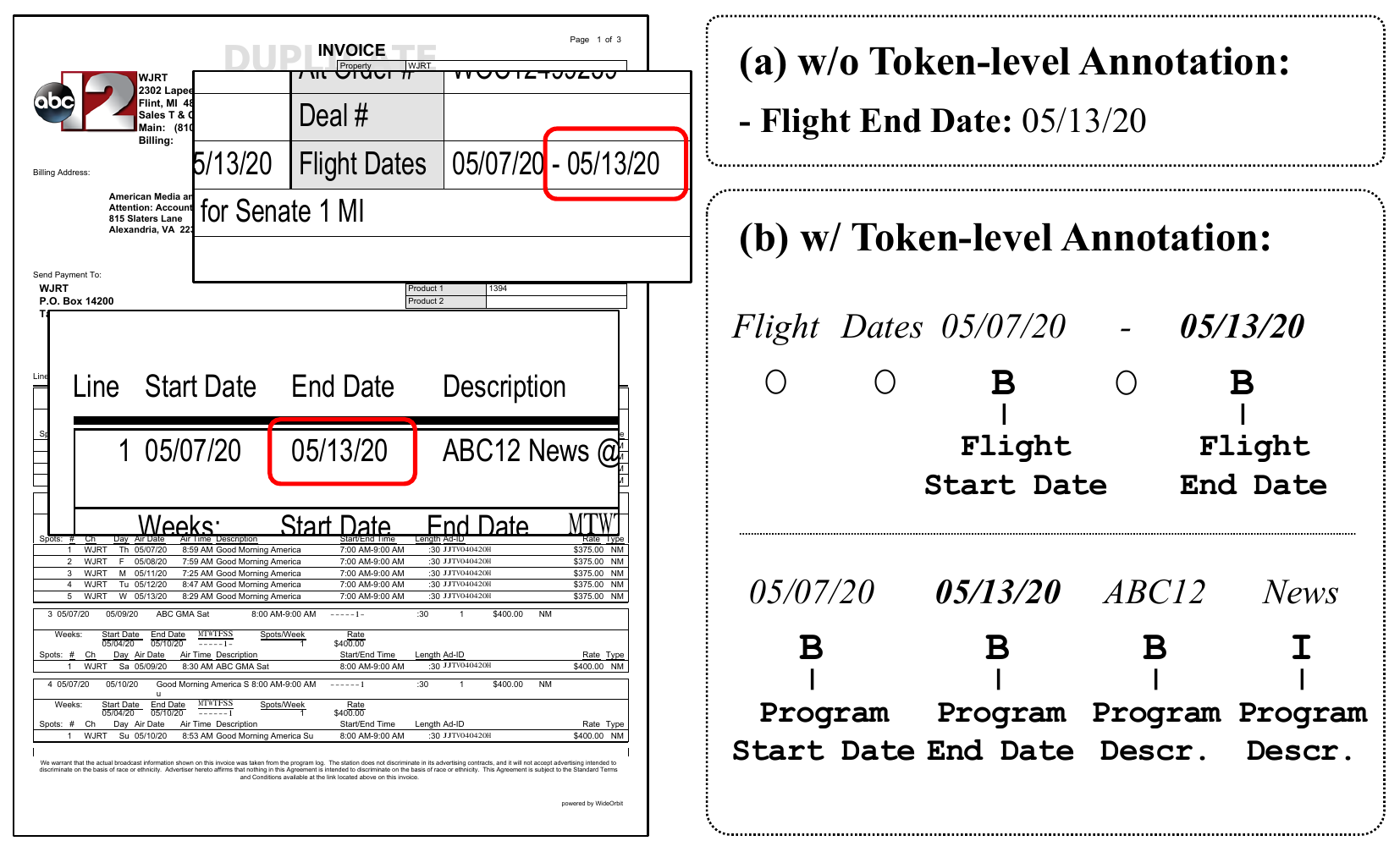}
    \caption{Examples of token-level annotation in visually-rich documents: (a) the dataset without token-level annotation where only the textual contents of entities are provided, and it is non-trivial to tell which ``05/13/20'' in the page is the value of \textit{flight end date}; (b) the dataset with token-level annotation where all tokens are labeled with BIO tags.}
    \label{fig:token_annotation}
\end{figure}

\subsection{Token-level Annotation}\label{sec:token-level}

A good benchmark ought to provide the token spans in the document that correspond to each entity in the target schema rather than simply provide text strings and leave the task of mapping the values to the corresponding token ranges open.
Existing approaches solve the extraction task using sequence labeling models and tend to build their models through extending BERT-like language models with multi-modal features~\cite{xu2020layoutlm,xu2020layoutlmv2,zhang2020trie,appalaraju2021docformer,garncarek2021lambert,huang2022layoutlmv3,wang2022lilt}. They use the hidden states from the language models to classify tokens into the BIO tags~\cite{marquez2005semantic,ratinov2009design}, i.e., \textit{Begin, Inside, Outside of an entity}, and then extract entities accordingly. Thus token spans are required to construct training and evaluation sets. It is non-trivial to re-construct the token-level annotation only with the entity text. The possible ways are either labor-intensive or prone to errors. A intuitive approach is to find the phrases in the documents with the same textual contents with the entities, but these phrases are not necessarily to be the actual entity, as shown in Figure \ref{fig:token_annotation}. \cite{tata2021glean} points out simply doing such value matching may result in worse F-1 scores in the performance. For instance, a dataset annotates the total amount field in a grocery store receipt as ``10'', but ``10'' may also appear in the receipt as the number of purchased items. Human annotators are needed to annotate the documents again to create the accurate token-level annotation. Therefore, token-level annotation is necessary to properly train and evaluate current baseline models and future works.



\section{Related Work}
\label{sec:relatedwork}
\begin{table*}[t]
  \centering
  \small
  \resizebox{\linewidth}{!}{
    \setlength{\tabcolsep}{2mm}{
\begin{tabular}{llccccccc}
    \toprule
    \multirow{3}[4]{*}{\textbf{Dataset}} & \multirow{3}[4]{*}{\textbf{Source}} & \multirow{3}[4]{*}{\textbf{Doc \#}} & \multirow{3}[4]{*}{\textbf{Entity \#}} & \multicolumn{5}{c}{\textbf{Desiderata}} \\
    \cmidrule{5-9}      &       &       &       & \textbf{Rich} & \textbf{Layout-rich} & \textbf{Diverse} & \textbf{High-quality} & \textbf{Token-level} \\
          &       &       &       & \textbf{Schema} & \textbf{Documents} & \textbf{Templates} & \textbf{OCR} & \textbf{Annotation} \\
    \midrule
    FUNSD            & Lawsuits Forms       & 199   & 3     & \xmark    & \cmark & \cmark & \cmark  & \cmark \\
    CORD             & Grocery Receipts     & 1000  & 30    & \cmark    & \cmark & \xmark & \xmark  & \cmark \\
    SROIE            & Grocery Receipts     & 973   & 4     & \xmark    & \cmark & \xmark & \xmark  & \xmark \\
    Kleister-NDA     & NDA Forms            & 540   & 4     & \xmark    & \xmark & \xmark & \cmark  & \xmark \\
    Kleister-Charity & Financial Reports    & 2778  & 8     & \xmark    & \cmark & \cmark & \cmark  & \xmark \\
    DeepForm         & FCC                  & 1100  & 5     & \xmark    & \cmark & \cmark & \cmark  & \xmark \\
    \midrule
    \textbf{VRDU-Registration Form} & \textbf{FARA} & \textbf{1915} & \textbf{6} & \Checkmark & \Checkmark & \Checkmark & \Checkmark & \Checkmark \\
    \textbf{VRDU-Ad-buy Form} & \textbf{FCC} & \textbf{641} & \textbf{9+1(5)$^*$} & \Checkmark & \Checkmark & \Checkmark & \Checkmark & \Checkmark \\
    \bottomrule
    \end{tabular}%
    }}
  \caption{The statistics of VRDU and other existing benchmarks. $^*$ denotes the number of hierarchical entities in the dataset, where VRDU-Ad-buy Form involves 1 hierarchical entity and the hierarchical entity has 5 entities as components.}
  \label{tab:statistics}%
\end{table*}%


Several benchmarks are available to evaluate the performance of models in visually-rich document understanding. The properties of these benchmarks and the comparison with our proposed benchmark are shown in Table \ref{tab:statistics}. 

FUNSD~\cite{jaume2019} is a dataset widely used in the form understanding task, which contains 199 fully annotated forms with three different entity types, \textit{Header}, \textit{Question}, and \textit{Answer}. This simple schema is too limited to reflect the rich schemas we encounter in practical scenarios. 
CORD~\cite{park2019cord} is a receipt dataset where the document images are photos of grocery receipts. While it does have a rich schema with different types including hierarchical and repeated fields, there is fairly limited template diversity. Furthermore, image artifacts (tilt, lighting, distortion) result in OCR errors. In our work, the focus is {\em not} on challenging OCR scenarios, but rather on benchmarks that help us understand how well models are able to extract information after high-quality OCR.
SROIE~\cite{huang2019icdar2019} is another receipt dataset. A few key fields are labeled, such as \textit{Company Name}, \textit{Address}, and \textit{Total Price} -- a fairly simple target schema. Further, the receipts in the dataset use the same template, failing to satisfy the requirement for diverse templates.
Kleister-NDA~\cite{stanislawek2021kleister} collects non-disclosure agreements and labels important fields but the documents are full of plain text paragraphs and chapters and show few layout elements. 


Kleister-Charity~\cite{stanislawek2021kleister} and DeepForm~\cite{svetlichnaya2020deepform,borchmann2021due} collect charity financial reports and political ad-buy documents respectively. Compared with the datasets above, DeepForm and Kleister-Charity involve layout-rich documents of various templates. However, both of them fail to provide token-level annotation. Further, both datasets have a target schema with multiple types, but lacking hierarchical and repeated fields.
As we describe in Section \ref{sec:token-level}, token-level annotations are critical to properly training and evaluating sequence labeling models.
Upon investigation, we found that the source documents for DeepForm do contain many more fields including hierarchical and repeated fields. We based one of the two dataset in VRDU, the ad-buy forms on the same source and designed the labeling task to include bounding boxes and token-level annotations.


This paper proposes VRDU, composed of two datasets of Registration Form and Ad-buy Form, both of which have rich schema, layout-rich documents, diverse templates, high-quality OCR outputs, and token-level annotations. The Ad-buy Forms provide hierarchical entity annotations, introducing a practical structural extraction task that has not been explored in any of the existing benchmarks.

\section{VRDU Benchmark}\label{sec:benchmark}
Based on the desiderata outlined in Section~\ref{section:desiderata}, we introduce VRDU, a new public benchmark for visually-rich document understanding. This benchmark includes two datasets: Ad-buy Forms and Registration Forms. These documents contain structured data with rich schema including hierarchical repeated fields, have complex layouts that clearly distinguish them from long text documents, have a mix of different templates, and have high-quality OCR results. We provide token-level annotations for the ground truth ensuring there is no ambiguity when mapping the annotations to the input text. In the remainder of this section, we describe: (1) the process used for collecting and annotating the datasets, (2) the three extraction tasks we designed along with the prescribed train/validation/test splits, and (3) the design and implementation of the type-aware matching algorithm used to compare the extracted entities with the ground-truth


\subsection{Data Collection}\label{sec:data}

Visually-rich documents are common in various business workflows. However, there are still a large proportion of documents that fail to meet our proposed desiderata. To make things worse, documents with sensitive information can only be used as in-house datasets due to privacy issues, so they are unsuitable for public academic research. 
To find visually-rich documents that satisfy our desiderata and are available to the public, we crawl political ad-buy forms from the same resource as the DeepForm dataset, the Federal Communications Commission, and construct a new dataset, the Ad-buy Forms. DeepForm includes documents of high quality but fails to provide token-level annotation with rich schema so we collect the documents from the same source and annotate them from scratch. We also crawl documents from the Foreign Agents Registration Act and construct a separate dataset, the Registration Form.
We use the state-of-the-art commercial OCR engines to recognize the raw data in the documents\footnote{\url{https://cloud.google.com/vision/docs/ocr}}.

\subsubsection{Ad-buy Forms}
The Ad-buy Forms consist of 641 documents about political advertisements. Each document is an invoice or receipt signed between a TV station and a campaign group. The documents use tables, multi-columns, and key-value pairs to record the advertisement information, such as the product name, the flight dates, and the total price. They also have a large table showing more details of the advertisements including the specific release date and time.



\subsubsection{Registration Forms}
The Registration Forms consist of 1915 documents about foreign agents registering with the US government. Each document records essential information about foreign agents involved in activities that require public disclosure. Contents include the name of the registrant, the address of related bureaus, the purpose of activities, and other details. We include three forms in the dataset, so the documents have three different templates, \textit{Amendment}, \textit{Short Form}, and \textit{Dissemination Report}. All these forms are on the same topic so we label them using the same schema. 
\begin{table}[t]
    \centering
    \small
       \resizebox{\linewidth}{!}{
      \setlength{\tabcolsep}{3mm}{
      \begin{tabular}{ll}
      \toprule
      \multicolumn{2}{c}{\textbf{Registration Form}} \\
      \midrule
      \multirow{2}[2]{*}{Unrepeated Entity} & \textit{file\_date, foreign\_principle\_name, registrant\_name,} \\
            & \textit{registration\_ID, signer\_name, signer\_title,} \\
      \bottomrule
            &  \\
      \toprule
      \multicolumn{2}{c}{\textbf{Ad-buy Form}} \\
      \midrule
      \multirow{2}[2]{*}{Unrepeated Entity} & \textit{advertiser, agency, contract\_ID, property, gross\_amount} \\
            & \textit{product, TV\_address, flight\_from\_date, flight\_to\_date} \\
      \midrule
      Repeated Entity & \textit{description, start\_date, end\_date, sub\_price} \\
      \midrule
      \multirow{2}[2]{*}{Hierarchical Entity} & \textit{line\_item} (composed of \textit{description, sub\_price, } \\
            & \textit{start\_date, end\_date}) \\
      \bottomrule
      \end{tabular}%
  }}
   \caption{The labeling schema of VRDU.}
    \label{tab:labeling_schema}%
  \end{table}%

\subsection{Human Annotation}

After we collect visually-rich documents for the two datasets, we hire human annotators to annotate entities in the documents using a rich labeling schema. We describe the labeling schema, the labeling team, and the label protocol as follows.

\subsubsection{Labeling Schema}

The documents in our proposed benchmark present structured data with fairly rich schema, where entities can be repeated, unrepeated, or hierarchical, and the data types can be numerical strings, price values, etc. After examining a subset of the documents, we decide the target schema with 6 unrepeated entity names for Registration Forms, and 9 unrepeated entity names and 1 hierarchical repeated entity name for Ad-buy Forms. The entity names and their numbers are shown in Table \ref{tab:labeling_schema} and Table \ref{tab:datset_stats}. 
\begin{itemize}[nosep, leftmargin=*]
    \item The unrepeated entities are the entities that only have one unique value in each document. Sometimes they may be present multiple times on a document, but with each instance having the exact same value. For example, a document may have several fields showing the contract ID but all these fields have the same content. 
    \item The repeated entities are the entities that belong to the same type but have different values. For example, the names of purchased items are common repeated entities in grocery receipts. People may buy several items so there will be multiple values for the entity type, \textit{purchased\_item\_name}.
    \item The hierarchical entities are the entities containing several repeated entities as components. For example, in Ad-buy Form, we design the \textit{line\_item} as a hierarchical entity, which corresponds to each TV program. Each \textit{line\_item} contains \textit{description}, \textit{start/end\_date}, and \textit{sub\_price} of TV programs and all of these are repeated entities. In practice, we group the repeated entities that belong to a specific TV program as a \textit{line\_item}. 
    
\end{itemize}




\subsubsection{Labeling Team}
We hired a labeling team of 30 annotators and 3 experts. All annotators and experts are experienced in labeling English documents and all of our data are in English. In our labeling task, the documents were first labeled by the annotators and then checked by the experts to guarantee the labeling quality. We acquired stats from our team of annotators on how long the classic annotation takes for various document types. We found it averaged 6-8 min for an annotator to label a single-page document with fewer than 20 fields while it averaged 10-30 min for an annotator to label a multi-page document with 25 fields. So we picked a conservative value (6 min) as the estimated time of labeling one document in this paper.

\begin{table}[htbp]
  \centering
  \small
   \resizebox{\linewidth}{!}{
  \setlength{\tabcolsep}{4mm}{
\begin{tabular}{cccc}
\toprule
\multicolumn{4}{c}{\textbf{Registration Form}} \\
\midrule
\textbf{Entity} & \textbf{Number} & \textbf{Entity} & \textbf{Number} \\
\midrule
Registration ID & 1903  & Foreign Principal & 1132 \\
Registrant Name & 1902  & Signer Name & 1467 \\
File Date & 1873  & Signer Title & 549 \\
\bottomrule
      &       &       &  \\
\toprule
\multicolumn{4}{c}{\textbf{Ad-buy Form}} \\
\midrule
\textbf{Entity} & \textbf{Number} & \textbf{Entity} & \textbf{Number} \\
\midrule
Property & 595   & Flight From Date & 540 \\
TV Address & 535   & Flight To Date & 538 \\
Advertiser & 635   & Gross Amount & 629 \\
Product & 607   & Agency & 283 \\
Contract ID & 624   & \textit{Line Item} & \textit{9163} \\
\bottomrule
\end{tabular}%

    }}
   \caption{The statistics of entity numbers in VRDU. The \textit{italic} entity names are hierarchical entities, which includes several repeat entities as components.}
    \label{tab:datset_stats}%
\end{table}%

\subsubsection{Labeling Protocol}

During the annotation, a pool of experienced annotators were provided with the previously annotated documents as reference and the labeling instruction as guidance. They drew bounding boxes to highlight the entities and labeled each entity into different categories. The system would collect the OCR results of the token span in the bounding box to construct token-level annotation, including the coordinates of the bounding box, the textual contents of entity, and the index in the sequence. If unrepeated entities occurred multiple times, they were instructed to identify all instances and the model only needs to extract one of them in the evaluation. When labeling the hierarchical entities, the annotators labeled the component entities as well as drew a larger bounding box that grouped the components together into a hierarchical entity. The system would use the entities in the larger box to compose hierarchical entities in our dataset. After the first pass of annotation, a pool of experts were assigned to review the results labeled by the first pool. 
We took the final corrected results from the expert pool and used them in our experiments. This is the dataset we published.

\subsubsection{Common Labeling Errors}
To better understand the labeling protocol, we further study the annotators’ common error types. 
\begin{itemize}[nosep, leftmargin=*]
\item Confusion of similar entities: In Ad-buy Form, the annotators are sometimes confused between the start/end dates of the flight and other time periods in the documents (e.g. the invoice period). \item Incomplete multi-line entities: In the Ad-buy Form dataset, the annotators sometimes ignore the last line of the address field since the address field usually contains multiple lines. 
\end{itemize}
To cope with these errors, we give annotators previous annotated documents as reference and ask another expert annotator to double check all the annotation results. We believe our labeling protocol can well prevent the annotation mistakes and produce a high-quality benchmark.


\subsection{Task Settings}\label{sec:task}

We design three tasks with increasing difficulty:

\subsubsection{Task 1: Single Template Learning (STL)}
This is the simplest scenario where the training, testing, and validation sets only contain a single template. This simple task is designed to evaluate a model's ability to deal with a fixed template. Naturally, we expect very high F1 scores for this task. 

\subsubsection{Task 2: Mixed Template Learning (MTL)}
This task is similar to the task that most related papers use: the training, testing, and validation sets all contain documents belonging to the same set of templates. We randomly sample documents from the datasets and construct the splits to make sure the distribution of the each template is not changed during the sampling.


\subsubsection{Task 3: Unseen Template Learning (UTL)}
This is the hardest setting, where we evaluate if the model is able to generalize to unseen templates. For example, in the Registration Forms dataset, we train the model with two of the three templates and test the model with the remaining one. The documents in the training, testing, and validation sets are drawn from disjoint sets of templates. To our knowledge, previous benchmarks and datasets do not explicitly provide such a task designed to evaluate the model's ability to generalize to templates not seen during training.

\subsubsection{Dataset Splits} In each of the task mentioned above, we include 300 documents in the testing set. We build 4 different training sets with 10, 50, 100, 200 samples respectively. The objective is to evaluate models on their data efficiency. The prescribed dataset splits are published along with the datasets to enable and apples-to-apples comparison between different models using this benchmark.


\subsection{Evaluation Toolkit}\label{sec:evaluation_method}
To evaluate extraction performance, we propose a type-aware fuzzy matching algorithm for each of the entities in our benchmark and report both the macro and micro F1 score for the dataset. 




\begin{algorithm}
  \footnotesize
   \caption{Entity Grouping}
    \begin{algorithmic}[1]
    \Function{Group}{$T$, $E$} 
    \Comment{$T$ is a set of entity names to be hierarchical, $E$ is an entity list.}
        \State $E^\prime = \{ e \in E | e.type \in T \}$ \Comment{$E^\prime$ includes all component entities.}
        \State $N = \phi$ \Comment{$N$ is to record all hierarchical entities.}
        \State $M = \phi$ \Comment{$M$ is to memorize entity names.}
        \State $i = 1, j = 1$
        \While{$i\le j \le E^\prime.length$}
            \If {$E^\prime[j].type \not\in M$}
                \State $M = M \cup \{E[j].type\}$
                \State $j = j + 1$
            \ElsIf {$E^\prime[j].type \in M$} \Comment{Group entities at repeated types}
                \State $N = N \cup \{E^\prime[i:j-1]\}$
                \State $i = j$
                \State $M = \phi$   \Comment{$M$ is reset to refresh memory.}
            \EndIf
        \EndWhile
        \State \Return $N$
    \EndFunction
\end{algorithmic}
\end{algorithm}
It is common practice to compare the extracted entity with the ground-truth using strict string matching~\cite{wolf2019huggingface}. However, such a simple approach may lead to unreasonable results in many scenarios. For example, ``\$ 40,000'' does not match with ``40,000'' because of the missing dollar sign when extracting the total price from a receipt, and ``July 1, 2022'' does not match with ``07/01/2022''. Dates may be present in different formats in different parts of the document, and a model should not be arbitrarily penalized for picking the wrong instance. We implement different matching functions for each entity name based on the data type. In the examples mentioned before, we will convert all price values into a numeric type before comparison. Similarly, date strings are parsed, and a standard date-equality function is used to determine equality.

\subsection{Post-processing for Evaluation Toolkit}\label{sec:post-processing}
We include repeated, unrepeated, or hierarchical entity names in our proposed VRDU benchmark. Our benchmark requires the model to predict a unique value for unrepeated entity names and group component entities into a hierarchical entity. However, such constraints are usually ignored by existing models. 
For example, the \textit{series ID} is an unrepeated entity and each document should only have one unique value for it, so the model is expected to extract a single string with the highest confidence instead of providing a number of candidates for the users to choose from. When there is no confidence score provided by the model, we simply keep the first extracted entity as the answer for the unrepeated entity names.

The hierarchical entity is a new kind of entity name proposed by our benchmark. Since existing works only focus on the extraction of individual entities, we propose a heuristic method to group the related individual entities into hierarchical ones and evaluate the result accordingly. The method is shown in Algorithm 1. Specifically, the repeated entities are first extracted from the document by the extraction model. Then, we list all these entities according to their index in the reading order extracted by the OCR engine. Then, we run our algorithm to split the list into several spans and each span corresponds to a hierarchical entity. The split point is decided by the occurrence of entity types. Briefly, when an entity type appears the second time, we split the list and build a hierarchical entity with the span. For example, supposing we have 3 entity types, A, B, and C, the extracted list, [A, B, C, B, C], would be divided into [A, B, C] and [B, C] where the split point is the second B in the list.



\begin{table*}[t]
      \centering
      \small
      \resizebox{\linewidth}{!}{
        \setlength{\tabcolsep}{3mm}{
    \begin{tabular}{clcccccccccc}
    \toprule
    \multirow{4}[6]{*}{$|\mathcal{D}|$} & \multirow{4}[6]{*}{\textbf{Model}} & \multicolumn{6}{c}{\textbf{Registration Form}} & \multicolumn{4}{c}{\textbf{Ad-buy Form}} \\
    \cmidrule(lr){3-8} \cmidrule(lr){9-12}   &       & \multicolumn{2}{c}{\textbf{Task 1}} & \multicolumn{2}{c}{\textbf{Task 2}} & \multicolumn{2}{c}{\textbf{Task 3}} & \multicolumn{2}{c}{\textbf{Task 2}} & \multicolumn{2}{c}{\textbf{Task 3}} \\
          &       & \multicolumn{2}{c}{\textbf{(Single Template)}} & \multicolumn{2}{c}{\textbf{(Mixed Template)}} & \multicolumn{2}{c}{\textbf{(Unseen Template)}} & \multicolumn{2}{c}{\textbf{(Mixed Template)}} & \multicolumn{2}{c}{\textbf{(Unseen Template)}} \\
    \cmidrule(lr){3-8} \cmidrule(lr){9-12}   &       & \textbf{Micro-F1} & \textbf{Macro-F1} & \textbf{Micro-F1} & \textbf{Macro-F1} & \textbf{Micro-F1} & \textbf{Macro-F1} & \textbf{Micro-F1} & \textbf{Macro-F1} & \textbf{Micro-F1} & \textbf{Macro-F1} \\
    \midrule
    \multirow{4}[2]{*}{\textbf{10}} & \textbf{LayoutLM} & 65.91 & 53.64 & 36.41 & 28.98 & 25.54 & 18.37 & 20.20 & 48.13 & 19.92 & 47.73 \\
          & \textbf{LayoutLMv2} & 80.05 & 68.89 & 69.44 & 63.79 & 54.21 & 45.38 & 25.36 & 58.13 & 25.17 & 57.84 \\
          & \textbf{LayoutLMv3} & 72.51 & 61.13 & 60.72 & 53.37 & 21.17 & 15.15 & 10.16 & 21.97 & 10.01 & 21.89 \\
          & \textbf{FormNet} & 74.22 & 62.95 & 63.61 & 56.53 & 50.53 & 40.24 & 20.47 & 55.15 & 20.28 & 54.80 \\
    \midrule
    \multirow{4}[2]{*}{\textbf{50}} & \textbf{LayoutLM} & 86.21 & 74.76 & 80.15 & 76.46 & 55.86 & 46.43 & 39.76 & 79.77 & 38.42 & 79.21 \\
          & \textbf{LayoutLMv2} & 88.68 & 77.51 & 84.13 & 82.04 & 61.36 & 52.42 & 42.23 & 83.89 & 41.59 & 84.14 \\
          & \textbf{LayoutLMv3} & 87.24 & 75.86 & 81.36 & 77.48 & 47.85 & 38.59 & 39.49 & 79.22 & 38.43 & 79.05 \\
          & \textbf{FormNet} & 89.38 & 78.04 & 85.38 & 82.41 & 68.29 & 57.17 & 40.68 & 83.82 & 39.52 & 83.49 \\
    \midrule
    \multirow{4}[2]{*}{\textbf{100}} & \textbf{LayoutLM} & 88.70 & 78.79 & 86.02 & 84.04 & 63.68 & 53.43 & 42.38 & 83.41 & 41.46 & 82.27 \\
          & \textbf{LayoutLMv2} & 90.45 & 80.03 & 88.36 & 86.38 & 65.96 & 57.39 & 44.97 & 86.38 & 44.35 & 85.62 \\
          & \textbf{LayoutLMv3} & 89.23 & 78.91 & 87.32 & 85.06 & 57.69 & 47.84 & 42.63 & 82.66 & 41.54 & 81.51 \\
          & \textbf{FormNet} & 90.91 & 80.82 & 88.13 & 85.82 & 72.58 & 62.23 & 40.38 & 84.24 & 39.88 & 83.57 \\
    \midrule
    \multirow{4}[2]{*}{\textbf{200}} & \textbf{LayoutLM} & 90.47 & 81.77 & 87.94 & 86.41 & 70.47 & 59.46 & 44.66 & 85.85 & 44.18 & 84.55 \\
          & \textbf{LayoutLMv2} & 91.41 & 83.12 & 89.19 & 87.65 & 72.03 & 62.14 & 46.54 & 87.61 & 46.31 & 86.87 \\
          & \textbf{LayoutLMv3} & 90.89 & 81.72 & 89.77 & 88.54 & 62.58 & 50.74 & 45.16 & 85.67 & 44.43 & 84.16 \\
          & \textbf{FormNet} & 92.12 & 82.99 & 90.51 & 89.05 & 77.29 & 67.82 & 43.23 & 86.08 & 42.87 & 85.05 \\
    \bottomrule
    \end{tabular}%
        }}
      \caption{Experiment results of Single Template Learning, Mixed Template Learning, Unseen Template Learning on Registration Form and Ad-buy Form.}
      \label{tab:result}%
    \end{table*}%

\section{Experiments}

We conduct experiments on VRDU and evaluate baseline models on the three proposed tasks. We report the micro-F1 and the macro-F1 scores across the training sizes proposed. 
Our primary goal with these experiments is to demonstrate that several challenges remain open in this space. In fact, while performance on other datasets discussed in Section~\ref{sec:relatedwork} might indicate that this is a solved problem, our results show all models fare worse on VRDU highlighting substantial room for improvements.
However, comprehensive comparison between existing models is an explicit {\em non-goal} for this paper.

\subsection{Baselines}
We evaluate three models on the datasets, LayoutLM~\cite{xu2020layoutlm}, LayoutLMv2~\cite{xu2020layoutlmv2}, LayoutLMv3~\cite{huang2022layoutlmv3}, and FormNet~\cite{lee2022formnet}.
\begin{itemize}[nosep, leftmargin=*]
    \item LayoutLM: LayoutLM is a layout-aware pre-trained language model which encodes the absolute coordinates of bounding boxes in the embedding layers of BERT~\cite{devlin2018bert} to inform the model of the structural information. Although the visual features from ResNet~\cite{he2016deep} are appended to the hidden states of LayoutLM to solve the task by the authors, we ignore them since they are not incorporated in the pre-training stage and only serve as add-on features to enhance performance. Thus, LayoutLM is a multi-modal language model with text and layout features.
    \item LayoutLMv2: LayoutLMv2 further improves the layout embedding in LayoutLM by considering the relative distance between different bounding boxes and proposes the two-stream multi-modal Transformer encoder to learn the correlation between the image and the text. The visual features are properly integrated in the Transformer framework, so LayoutLMv2 is a multi-modal language model with text, layout, and visual features.
    \item LayoutLMv3: LayoutLMv3 improves the modeling with image features. Cross-modality pre-training tasks are also incorporated to enhance the performance.
    \item FormNet: FormNet first uses the attention mechanism to model the 2D spatial relationship between words and further goes beyond simply sequence labeling approach by leveraging the graphs constructed by the layout elements in the documents to aggregate semantically meaningful information from neighboring tokens.
\end{itemize}
Although we acknowledge there are many other approaches to solving structured extractions from such documents~\cite{appalaraju2021docformer,biten2022latr,garncarek2021lambert,powalski2021going,wang2022lilt,zhang2020trie,lee2021rope,huang2022layoutlmv3}, we only consider these three commonly-used ones to highlight the challenges common to all three models and inspire possible directions for future study. As we said previously, a comprehensive comparison is outside the scope of this paper.

\begin{figure*}[t]
    \centering
    \includegraphics[width=0.8\linewidth]{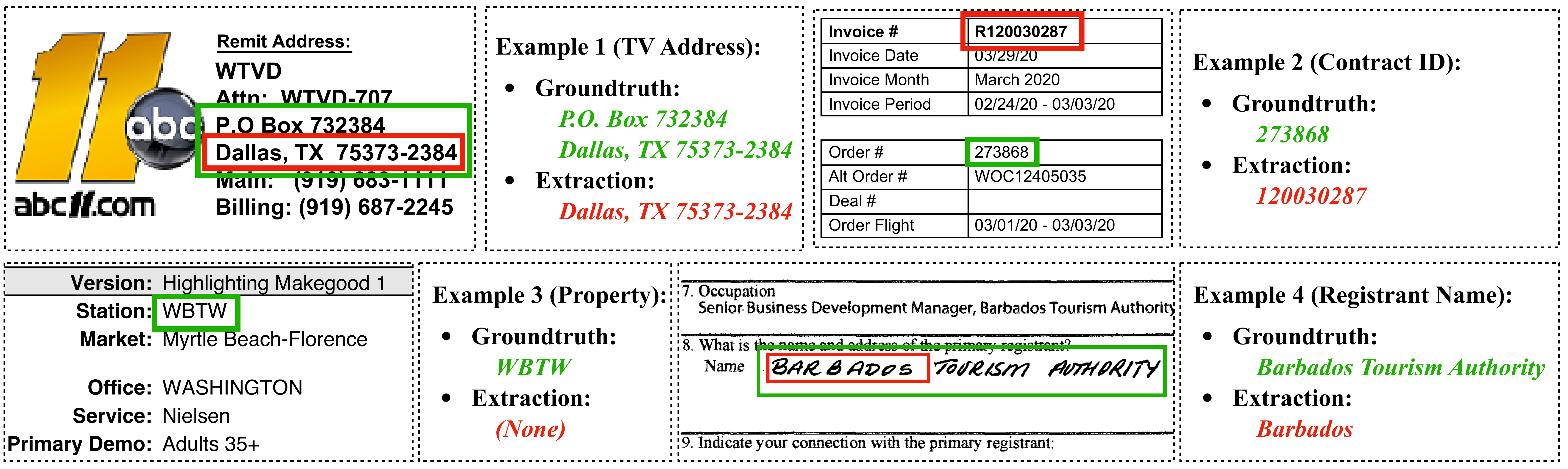}
    \caption{Loss cases found in the experiments: Example 1, 2, 3 are from Ad-buy Form, and Example 4 are from Registration Form. In each case, green indicates the ground-truth, and red indicates the extraction from the model.}
    \label{fig:case_study}
\end{figure*}

\begin{figure}
    \centering
    \includegraphics[width=0.75\linewidth]{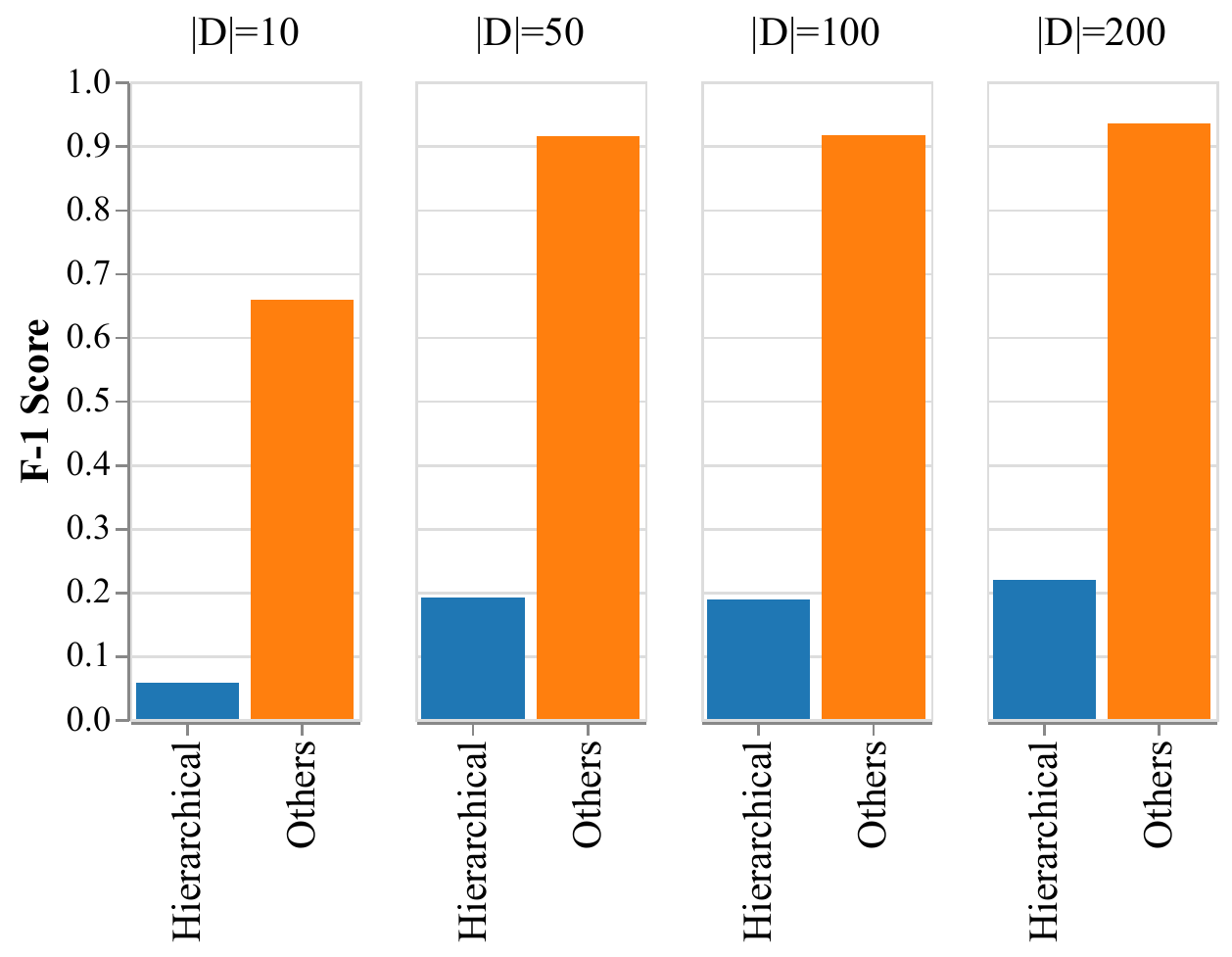}
    \caption{Comparison of FormNet on hierarchical and other entities in Mixed Template Learning, where $|\mathcal{D}|$ denotes the number of training samples.}
    \label{fig:line_item_performance}
\end{figure}

\subsection{Experiment Results}

We report the micro-F1 score and macro-F1 score of the three tasks, Single Template Learning (STL), Mixed Template Learning (MTL), and Unseen Template Learning (UTL), under different number of training samples in Table \ref{tab:result}. Since Ad-buy Form dataset contains a variety of templates and there are only a limited number of documents for each template, we skip the STL task for it.
We denote the number of training samples as $|\mathcal{D}|$. Under each setting, we build three training sets of the same size using different random seeds, and the reported numbers are the average result of each model on the three training sets. 

First, comparing the results on VRDU and on other benchmarks in Table \ref{tab:result}, it is clear that there is ample room for improvement. Even when $|\mathcal{D}|=200$, the highest micro-F1 score is around 90\% on Registration Form and around 45\% on Ad-buy Form. In contrast, FormNet achieves 97.21\% micro-F1 score and LayoutLMv2 achieves 96.01\% micro-F1 score on CORD~\cite{xu2020layoutlmv2,lee2022formnet}. LayoutLMv2 achieves 97.81\% micro-F1 score on SROIE~\cite{xu2020layoutlmv2}. One might think that results on CORD and SROIE indicate that this is a solved problem. As results on VRDU show, a dataset that reflects challenges in practical settings shows that there is much room for improvement. The performance of FormNet on FUNSD is 84.69\% micro-F1 score, and that of LayoutLmv2 is 84.20\% micro-F1 score~\cite{xu2020layoutlmv2,lee2022formnet}. Although there is still room to improve, the simplistic labeling schema used in FUNSD makes the results less representative of practical tasks.


We also observe consistent improvement as training data size increases. Even for the simplest task, STL (on Registration Forms), the micro-F1 score of FormNet when $|\mathcal{D}|=10$ is lower than that when $|\mathcal{D}|=50$ by 15.16 points. This 15+ point gap remains across all tasks for both datasets between the $|\mathcal{D}|=10$ and $|\mathcal{D}|=50$ settings. This holds true for all three models, underscoring that few-shot performance is difficult for all models, even for the simple STL setting getting to micro-F1 scores of just 74.22\%.





We then compare the performance of different tasks, STL, MTL, and UTL. The tasks are designed to study the template generalization of each model. From the results, we can see all models performs well in STL and MTL and achieve micro-F1 and macro-F1 scores higher than 80\% in both datasets with 200 training samples. We attribute the performance to the fact that there are no unseen layout structures involved when generalizing to the testing set in STL and MTL. However, there is a noticeable gap between the performance of MTL and UTL. At 200 training documents, micro-F1 for UTL is 13--17 percentage points worse than the micro-F1 for MTL across the three models. The performance of UTL on Ad-buy Form is worse than MTL by about 3 points. Recall that the test set in UTL contains documents with templates (layouts) not seen in the training set. We believe techniques that allow models to generalize to new layouts even with modest training sets are of practical importance.

Studying the performance in Ad-buy Form, we see the macro-F1 scores are much higher than the micro-F1 scores. The micro-F1 score weighs every instance of an entity equally, while the macro-F1 scores average the F1 score for each entity.
The huge difference between these scores for Ad-buy Form is because of the presence of hierarchical repeated entities with a very low F1 score. 





\subsection{Performance on Hierarchical Entities}\label{sec:nested_entity}

We next study the performance of hierarchical entities in Ad-buy Form dataset. Consider the performance of FormNet on MTL. The performance of extracting hierarchical entities vs. other entities is plotted in Figure \ref{fig:line_item_performance}. As we can see, there is a huge gap of 60 -- 70 points across different sizes of training sets when comparing the micro-F1 score of hierarchical entities and other entities. In contrast to unrepeated entities, the hierarchical entity requires the model not only to correctly extract the corresponding entities, but also to group the components together. Currently, a heuristic method is used as a simple baseline to deal with the hierarchical entity since no existing models take the hierarchical entity type into consideration. We describe the method in detail in Section \ref{sec:post-processing}. However, such a heuristic results in very low F1 scores for the entity. It is still an open question for future research how to properly extract the hierarchical entities from visually-rich documents.

\subsection{Case Study}
We select four loss cases in the experiments of FormNet and visualize the errors in Figure \ref{fig:case_study}.
We hope this spurs ideas for future improvements.

\subsubsection{Incomplete Extraction} Example 1 and 4 suffer from the incomplete extraction, i.e., the model can correctly locate the ground-truth entity but fails to include all the necessary information. In Example 1, the \textit{TV\_address} field is hidden in complex context, which makes it hard to recognize the P.O. Box as part of the address. In Example 4, the error of \textit{Registrant\_name} is because of the handwritten characters in different sizes and fonts. The models cannot group the characters together to extract the right entity. 

\subsubsection{Misleading Key Words} The errors in Example 2 and 3 result from misleading key words. Specifically, in Example 2, the model is confused by the similar key word, ``Invoice \#'', and extract the Invoice ID instead of the Order ID, although there are cases in the training set where the key word for \textit{contract\_ID} field is ``Order \#''. In Example 3, the model fails to extract any entity as \textit{Property} since the document is in a new template where ``Station'' is used as the key word for \textit{Property} field. To solve the rare case in Example 3, it is useful to take into consideration that ``WBTW'' is common in the training set as \textit{Property} field.

\section{Conclusions and Future Study}
In this paper, we identify five benchmark desiderata to measure progress on solving structured extractions from visually-rich documents in real application. We argue that existing benchmarks fall short on these and propose a new comprehensive benchmark, VRDU, including the dataset with high-quality OCR results and annotations, the tasks corresponding to different application scenarios, and the evaluation toolkit using the type-aware matching algorithm. Based on the novel task settings and extensive experiments, we highlight three areas of opportunity in the visually-rich document understanding task, including the generalization to new templates, the extraction under few-shot scenarios, and the extraction of complex hierarchical-repeated fields. We make the two datasets, all train/validation/test splits, and the evaluation toolkit publicly available. We hope this facilitates progress in this area.
In future study, we would further evaluate the existing models using our benchmark to understand how models perform when incorporating multi-modal features into the language models and explore whether there are any potential directions of new frameworks solving the task. We will also focus on the three areas of opportunity discovered in this paper, and explore approaches that can solve the visually-rich document understanding task in the scenarios with unknown templates, limited number of training samples, and hierarchical entities.

\section{Acknowledgement}
This work was supported by Google. We thank anonymous reviewers and program chairs for their valuable and insightful feedback. We thank Chun-Liang Li and Guolong Su from Google for their generous help in designing and running experiments.

\bibliographystyle{ACM-Reference-Format}
\balance
\bibliography{custom}



\end{document}